\documentclass[runningheads]{llncs}
\usepackage[T1]{fontenc}
\usepackage{graphicx}
\usepackage[misc]{ifsym} 
\usepackage{multirow}
\usepackage{bbding}
\usepackage{gensymb}
\usepackage{color}
\usepackage[colorlinks,linkcolor=blue]{hyperref}
\usepackage{amsmath}
\newcommand{\repeatthanks}{\textsuperscript{\thefootnote}}

\begin{document}
\title{Fine-grained Context and Multi-modal Alignment for Freehand 3D Ultrasound Reconstruction}

\author{
Zhongnuo Yan\inst{1,2,3}\thanks{Zhongnuo Yan and Xin Yang contribute equally to this work.} \and
Xin Yang\inst{1,2,3}\repeatthanks \and
Mingyuan Luo\inst{1,2,3} \and
Jiongquan Chen\inst{1,2,3} \and
Rusi Chen\inst{1,2,3} \and
Lian Liu\inst{1,2,3,4} \and
Dong Ni\inst{1,2,3}\textsuperscript{(\Letter)}
}

\authorrunning{Z. Yan et al.}
\titlerunning{Fine-grained Context and Multi-modal Alignment for Freehand 3D US}

%

\institute{
National-Regional Key Technology Engineering Laboratory for Medical Ultrasound, School of Biomedical Engineering, Shenzhen University Medical School, Shenzhen University, China \\
\email{nidong@szu.edu.cn} \and
Medical Ultrasound Image Computing (MUSIC) Lab, Shenzhen University, China \and
Marshall Laboratory of Biomedical Engineering, Shenzhen University, China \and
Shenzhen RayShape Medical Technology Inc., Shenzhen, China
}

\maketitle              
\begin{abstract}
Fine-grained spatio-temporal learning is crucial for freehand 3D ultrasound reconstruction. Previous works mainly resorted to the coarse-grained spatial features and the separated temporal dependency learning and struggles for fine-grained spatio-temporal learning. Mining spatio-temporal information in fine-grained scales is extremely challenging due to learning difficulties in long-range dependencies. In this context, we propose a novel method to exploit the long-range dependency management capabilities of the state space model (SSM) to address the above challenge. Our contribution is three-fold. First, we propose ReMamba, which mines multi-scale spatio-temporal information by devising a multi-directional SSM. Second, we propose an adaptive fusion strategy that introduces multiple inertial measurement units as auxiliary temporal information to enhance spatio-temporal perception. Last, we design an online alignment strategy that encodes the temporal information as pseudo labels for multi-modal alignment to further improve reconstruction performance. Extensive experimental validations on two large-scale datasets show remarkable improvement from our method over competitors.

\end{abstract}

\section{Introduction}
Freehand 3D ultrasound (US) can provide comprehensive spatial information about the scanned region of interest and has been widely used in clinical diagnosis~\cite{mohamed2019survey,prevost20183d}.
With the development of deep learning technology, current freehand 3D ultrasound reconstruction is free from dependence on external positioning devices, which were previously routinely utilized. They reconstruct the volume by estimating the relative spatial transformations of a series of US images. 
However, the difficulty in mining spatio-temporal information in fine-grained scales makes it very challenging to accurately infer the relative position.

Recent studies were mainly based on convolutional neural network (CNN) and achieved advanced performance. Prevost et al.~\cite{prevost2017deep} introduced an end-to-end method utilizing CNN to estimate the relative motion of US images. Guo et al.~\cite{guo2022ultrasound} proposed a deep contextual-contrastive network (DC$^{2}$-Net) and introduced a contrastive learning strategy to enhance reconstruction performance. Li et al.~\cite{li2023trackerless} proposed to estimate 3D spatial transformation between US frames using recurrent neural networks (RNNs). Luo et al.~\cite{luo2021self,luo2023recon} further improves reconstruction performance by online learning and shape priors. However, we note that the general approach of these studies is to first extract the coarse-grained features of the image and then extract the temporal information contained in these features. This design undoubtedly ignores the fine-grained spatio-temporal information, which is crucial for freehand 3D ultrasound reconstruction, and results in fragmentation between spatial and temporal information. 

\begin{figure}[tbp]
\centering
\includegraphics[width=\textwidth]{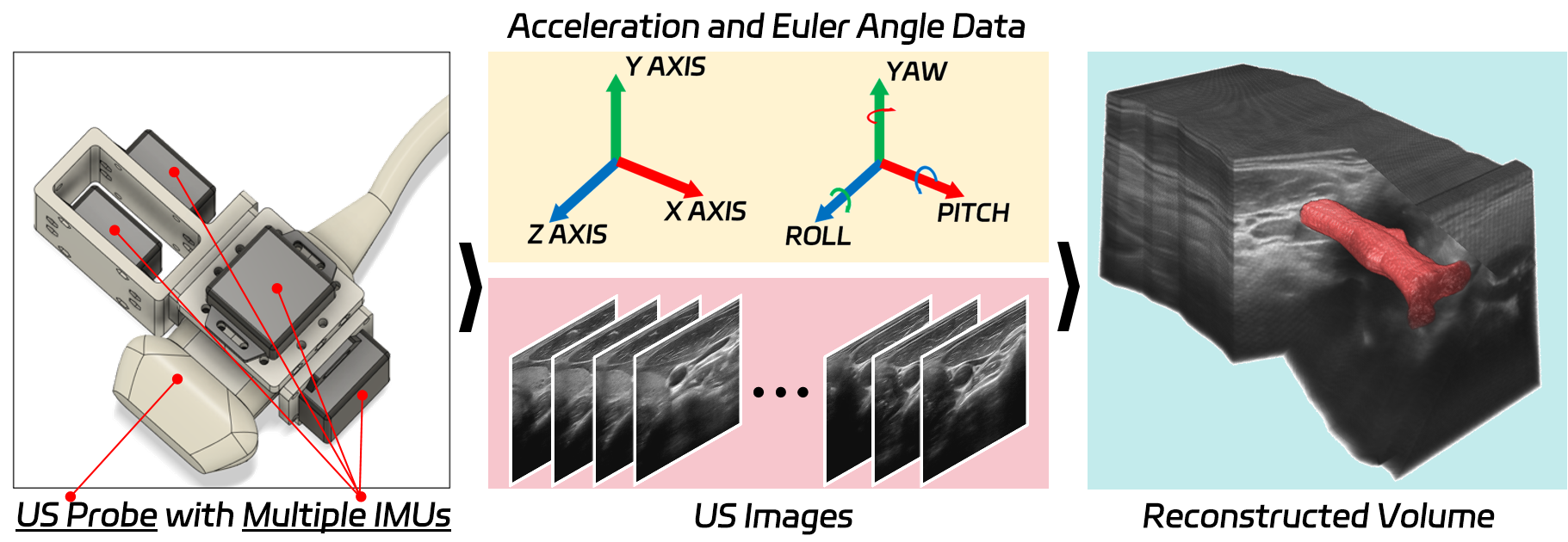}
\caption{Pipeline of freehand 3D US reconstruction with multiple IMUs.} 
\label{fig:intro}
\end{figure}

The lightweight sensor known as the inertial measurement unit (IMU) is an ideal choice for freehand 3D ultrasound reconstruction due to its low cost, low power consumption, and small size, as shown in Figure~\ref{fig:intro}. Prevost et al.~\cite{prevost20183d} have shown that incorporating IMU angles can enhance the accuracy of relative motion estimation. Luo et al.~\cite{luo2022deep,luo2023multi} have developed two multi-modal networks that leverage the valuable information from acceleration and angle measurements obtained from single or multiple IMUs to improve reconstruction performance. These studies highlight the significant improvement that IMUs can bring to freehand 3D ultrasound reconstruction.

In this study, we propose FiMA (\textbf{Fi}ne-grained Context and \textbf{M}ulti-modal \textbf{A}lignment), which exploits the efficient long-range dependency management capabilities of the state space model (SSM) to mine the spatio-temporal information in fine-grained features. Our contribution primarily revolves around three key aspects. First, we propose ReMamba, which mines multi-scale spatio-temporal information via multi-directional SSM. Additionally, we propose an adaptive fusion strategy that introduces multiple IMUs as additional temporal information to enhance spatio-temporal information. Last, we design an online alignment strategy that uses the temporal information of IMUs as pseudo-labels for multi-modal alignment to further improve reconstruction performance.

\section{Methods}
\begin{figure}[tbp]
\centering
\includegraphics[width=\textwidth]{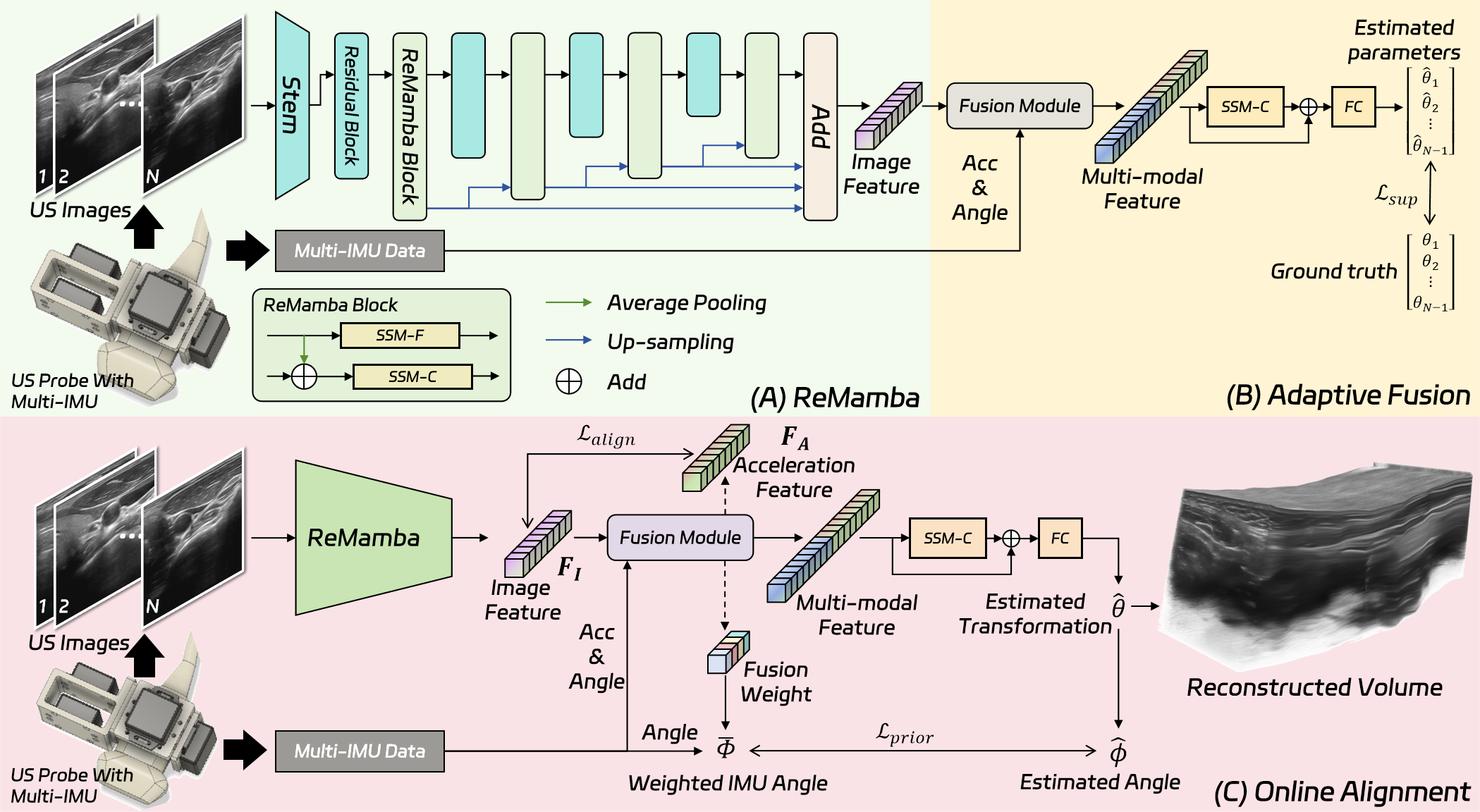}
\caption{Overview of the proposed FiMA.} 
\label{fig:main-arch}
\end{figure}

An overview of our proposed FiMA is shown in Figure~\ref{fig:main-arch}. It consists of three components: ReMamba for image sequence encoding (Figure~\ref{fig:main-arch}(A)), adaptive fusion (Figure~\ref{fig:main-arch}(B)) and online alignment (Figure~\ref{fig:main-arch}(C)). Given an $N$-length scanning sequence $I=\{I_i|i=1,2,\cdots,N\}$ and corresponding multiple IMU data $U=\{U_i|i=1,2,\cdots,N-1\}$, we utilize FiMA to estimate the transformation parameters $\theta=\{\theta_i|i=1,2,\cdots,N-1\}$. In this context, $\theta_i$ refers to the 3-axis translations $t_i=(t_x,t_y,t_z)_i$ and rotation angles $\phi_i=(\phi_x,\phi_y,\phi_z)_i$ between image $I_i$ and $I_{i+1}$. There are $M$ independent IMU data $U_i=\{U^j_i|j=1,2,\cdots,M\}$. Here, $U^j_i$ consists of 3-axis angles $\Phi^j_i=(\Phi_x,\Phi_y,\Phi_z)^j_i$ and accelerations $A^j_i=(A_x,A_y,A_z)^j_i$. The pre-processing process for $\Phi_i$ and $A_i$ follows the method described in~\cite{luo2022deep}.

\subsection{ReMamba with Multi-directional State Space Model}
Fine-grained spatio-temporal information is crucial for accurate reconstruction. Previous methods mainly capture temporal information just in coarse-grained features, which is typically due to mining spatio-temporal information in fine-grained features involves intractable long-range dependencie. Recently, Mamba~\cite{gu2023mamba}, with a SSM architecture and hardware-aware algorithms, demonstrated excellent long sequence processing capabilities. Inspired by SSM, we propose ReMamba, which mines multi-scale spatio-temporal information via multi-directional SSM. ReMamba consists of stem, multiple residual blocks and ReMamba blocks with multi-directional SSM. The architecture of ReMamba is shown in Figure~\ref{fig:main-arch}(A).

\textbf{Preliminaries.} SSM maps a sequence $x(t)$ to another sequence $y(t)$ through a implicit latent state $h(t)$. It contains three learnable matrices $\mathbf{A}$, $\mathbf{B}$, $\mathbf{C}$ and satisfies the following system of equation:
\begin{equation}
\begin{aligned}
h^\prime(t)&=\mathbf{A}h(t)+\mathbf{B}x(t) \\
y(t)&=\mathbf{C}h(t)
\end{aligned}
\end{equation}

The system is continuous, SSM is less effective on discrete data such as image. Mamba is a discrete version of continuous system, it makes SSM parameters a function of the input, which allows the model to selectively propagate or forget information based on current token. This facilitates compression of the context into a small state and better management of long-range dependency.

\textbf{ReMamba Block.}
We flatten the three-dimensional image sequence features into one-dimensional sequences in order to model spatial and temporal information in a unified perspective, capturing temporal and spatial information in multi-scale features. Different flatten rules produce different contexts. To capture spatio-temporal information that is suitable for reconstruction, we design two different granularities of SSM within each ReMamba block, fine-grained \textbf{SSM-F} and coarse-grained \textbf{SSM-C}. Each ReMamba block takes as input two features of different scales and outputs two features of the same shape.

\begin{figure}[tbp]
\centering
\includegraphics[width=\textwidth]{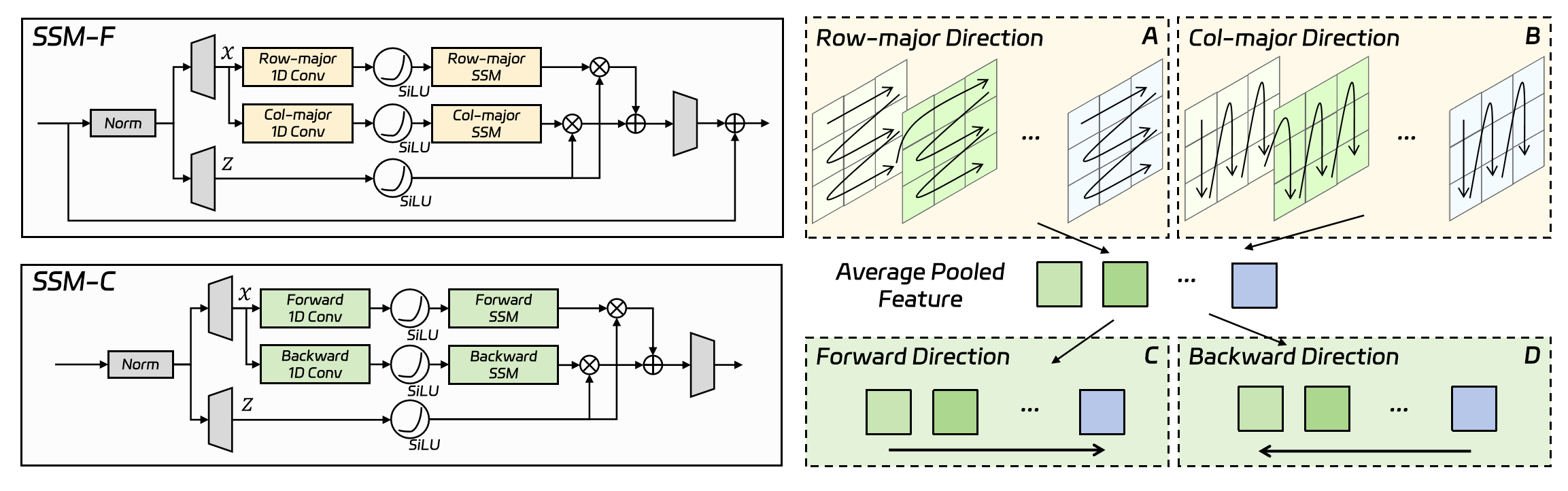}
\caption{Detail design of ReMamba Block. } \label{fig:remamba-block}
\end{figure}

SSM-F is used to capture diverse spatio-temporal information in multi-scale features. SSM-F receives output of previous residual block as input, and after normalised distribution by LayerNorm, it goes through linear projections on the main branch and on the gated branch into high-dimensional space to obtain $x$, $z$, respectively. $x$ then passes through the 1D convolution, SiLU and SSM layers in both row-major direction (Figure~\ref{fig:remamba-block}(A)) and col-major direction (Figure~\ref{fig:remamba-block}(B)). $z$ goes through SiLU and products with the outputs of SSM in two directions in the main branch, the products are summed up and finally linearly projected back to the original dimensions. The projection results are summed with the inputs to get fine-grained temporal information as output.

SSM-C is used to perceive overall motion patterns, the average pooling result of the previous residual block output is then summed with the previous SSM-C output, and the result as input to SSM-C, the input does not need to be flattened. Similarly, we use both forward (Figure~\ref{fig:remamba-block}(C)) and backward (Figure~\ref{fig:remamba-block}(D)) direction and finally get the coarse-grained temporal information as output.

\subsection{Adaptive Fusion Strategy}


\begin{figure}[tbp]
\centering
\includegraphics[width=\textwidth]{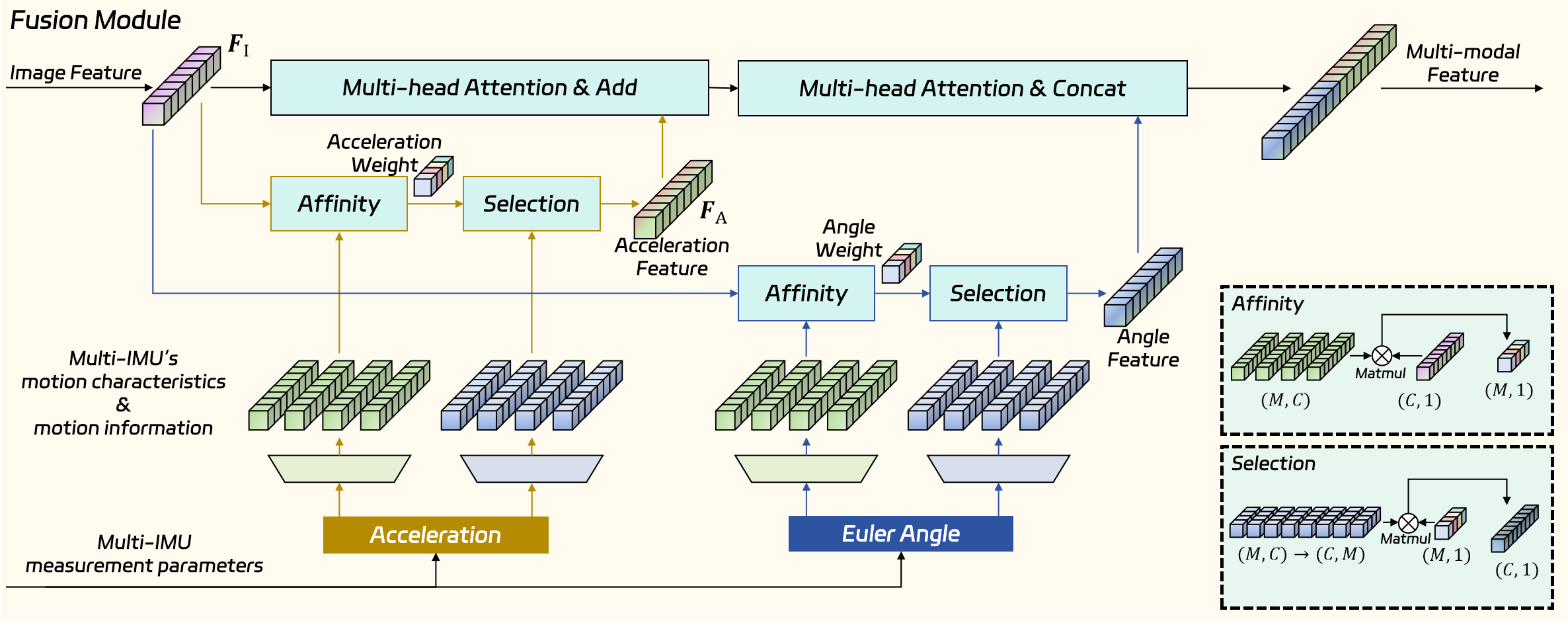}
\caption{Details of Fusion Module. Its input are the image features from ReMamba, the acceleration and Euler angles of multiple IMUs. It outputs multi-modal fused feature.}
\label{fig:fusion-module}
\end{figure}
IMU can provide motion information beyond the image, recording the state of motion over a period of time. To further enhance the spatio-temporal information, we introduce multiple IMUs, and consider the measurement parameters provided by IMUs as additional temporal information. However, motion of scan sequences is complex and various. In addition, due to each IMU having different characteristics and states and generating different noise, the temporal information provided by multiple IMUs is not always consistent.

We introduce an adaptive fusion strategy (Fusion Module in Figure~\ref{fig:main-arch} to address the above problem. Spatio-temporal information extracted from images and temporal information extracted from IMUs are not independent, instead, there should be a correlation between them. By combining the spatio-temporal information extracted from images, we expect the network to be able to judge whether each IMU provides reasonable temporal information at each moment. Specifically, as shown in Figure~\ref{fig:fusion-module}, we map the acceleration/angle of each IMU to the same two spaces as the image, representing the motion characteristics and detailed temporal information of this IMU, respectively. We calculate affinity between spatio-temporal information of images and representation for the motion characteristics of the acceleration/angle of each IMU, the affinity is used as a reference for weighting the detailed temporal information.

For the temporal information obtained from IMU acceleration/angle, we use two multi-head attention to fuse spatio-temporal information of images and temporal information of IMUs, respectively, and get the multi-modal fused features. Finally, we use a SSM-C module with skip connection and a linear projection to decode the estimated transformation parameters $\hat{\theta}$. In training phase, we optimise the network using mean absolute error and Pearson correlation loss:
\begin{equation}
\begin{aligned}
\mathcal{L}_{\text{sup}}=\Vert\hat{\theta}-\theta\Vert_1+\left(1-\frac{\textbf{Cov}(\hat{\theta}, \theta)}{\sigma(\hat{\theta})\sigma(\theta)}\right)
\end{aligned}
\end{equation}
where $\theta$ denotes the true transformation parameters, $\textbf{Cov}$ denotes the covariance calculation, and $\sigma$ denotes the standard deviation.

\subsection{Online Alignment Strategy}
To capture appropriate temporal features on unseen data, we propose an online alignment strategy (shown in Figure~\ref{fig:main-arch}(C)), which takes the multiple IMU information as pseudo-labels and further enhance the reconstruction performance through multi-modal feature alignment in test phase. 

For the image feature $F_I$ from ReMamba, and the weighted temporal feature of IMU acceleration $F_A$ from Fusion Module, we maximize the mutual information between them to align their feature spaces, while facilitating their mutual reduction of each other's uncertainty. However, it is difficult to optimize the mutual information directly, and we optimize one of its lower bounds refer to~\cite{oord2019representation}. For an N-length sequence, the alignment loss $\mathcal{L}_{\text{align}}$ is calculated according to the image feature $F_{Ii}$ and acceleration temporal feature $F_{Ai}$ ($0 < i < N$):
\begin{equation}
\begin{aligned}
\mathcal{L}_{\text{align}}=-\frac{1}{N-1}\sum_{i=1}^{N-1}\log\left(\frac{\exp(F_{Ii}\cdot F_{Ai}/\tau)}{\sum_{j=1}^{N-1}\exp(F_{Ii}\cdot F_{Aj}/\tau)}\right)
\end{aligned}
\end{equation}
where $\tau$ is the temperature parameter, and we set it to 0.1.

IMUs provide accurate angle measurements in most cases, we use the fusion weights corresponding to IMU angle in the fusion module as prior, and use the prior weight to calculate average angle $\Bar{\Phi}$ of these angle measurements, we calculate the loss between the estimated Euler angle $\hat{\phi}$ and weighted IMU angle using Pearson correlation loss: 
\begin{equation}
\begin{aligned}
\mathcal{L}_{\text{prior}}=1-\frac{\textbf{Cov}(\hat{\phi}, \Bar{\Phi})}{\sigma(\hat{\phi})\sigma(\Bar{\Phi})}
\end{aligned}
\end{equation}
We optimize FiMA using the sum of the above loss functions.


\section{Experiments}
\subsubsection{Materials and Implementation.} 
We construct two datasets refering to \cite{luo2023multi}, including arm and carotid, from 50 volunteers. The arm dataset contains 583 scans, employing a variety of scanning tactics such as linear, curved, loop, and sector scans. Similarly, the carotid dataset includes 432 scans, utilizing linear, loop, and sector scan tactics. The average lengths of the arm and carotid scans are $386.39$ and $241.25$ mm, respectively. The size of scanned images is $248 \times 260 $ pixels, with an image spacing of $ 0.15 \times 0.15$ mm$^{2}$. The positions of IMUs are the same as in Figure~\ref{fig:intro}. The procurement and application of this data received approval from the local Institutional Review Board (IRB), ensuring compliance with ethical guidelines. 

\begin{table}[!t]
\centering
\caption{The mean (std) results of different methods on the arm and carotid scans. *indicates that the method does not require a sensor. ReM$'$ and ReM use Mamba and ReMamba block, respectively. F$'$ and F represent the direct mapping of acceleration/angle of multiple IMU to high dimension combined and the proposed adaptive fusion strategy, respectively. The best results are shown in blue.}\label{tab1}
\begin{tabular}{c|cccccc}
\hline
  Method & FDR(\%)$\downarrow$ & ADR(\%)$\downarrow$ & MD(mm)$\downarrow$  & SD(mm)$\downarrow$ & HD(mm)$\downarrow$ & MEA(deg)$\downarrow$ \\
\hline
& \multicolumn{6}{c}{Arm dataset}\\
\hline
 CNN-OF* & 33.03(21.3) & 46.80(36.5) & 82.94(36.9) & 2360.97(1503.3) & 74.85(36.4) & 4.82(3.0)\\
 ResNet* & 21.13(12.8) & 31.29(18.0) & 55.01(28.2) & 1684.58(1795.8) & 51.10(26.9) & 7.36(4.2)\\
 DC$^{2}$-Net* & 18.12(12.7) & 26.63(15.0) & 48.26(30.2) & 1399.03(1670.2) & 48.15(30.8) & 7.02(4.0)\\
 RecON* & 15.37(9.7) & 22.23(11.7) & 34.41(20.1) & 1096.15(963.0) & 30.97(15.8) & 5.23(3.5)\\
 MoNet & 14.38(8.7) & 21.20(10.4) & 32.36(18.7) & 1009.60(863.5) & 28.96(14.2) & 3.70(2.3)\\
 OSCNet & 13.06(7.4) & 19.90(11.2) & 30.81(17.3) & 947.06(716.6) & 27.69(13.2) & 3.45(2.2))\\
\cline{1-7}
 ReM$'$ & 17.31(14.7) & 26.36(16.6) & 45.22(30.9) & 1236.33(1135.5) & 40.50(28.4) & 6.50(3.3)\\
 ReM & \textcolor{black}{13.87(12.5)} & \textcolor{black}{20.86(14.6)} & \textcolor{black}{35.07(26.0)} & \textcolor{black}{969.46(911.6)} & \textcolor{black}{32.42(24.9)} & \textcolor{black}{5.67(2.9)}\\
 ReM+F$'$ & 12.87(12.6) & 18.98(13.3) & 31.88(24.4) & 875.87(880.0) & 29.78(22.9) & 5.17(2.8)\\
 ReM+F & \textcolor{black}{10.85(8.0)} & \textcolor{black}{16.86(10.4)} & \textcolor{black}{27.38(16.1)} & \textcolor{black}{746.32(567.2)} & \textcolor{black}{25.60(15.9)} & \textcolor{black}{4.58(2.6)}\\
 FiMA & \textcolor{blue}{9.72(7.1)} & \textcolor{blue}{15.53(9.6)}& \textcolor{blue}{24.68(13.6)} & \textcolor{blue}{677.48(498.4)} 
& \textcolor{blue}{23.05(13.6)}  & \textcolor{blue}{3.41(1.8)}\\
\hline
& \multicolumn{6}{c}{Carotid dataset}\\
\hline
 CNN-OF* & 28.25(18.3) & 42.87(21.6) & 45.12(17.5) & 1392.58(1057.0) & 39.68(16.6) & 3.95(2.9)\\
 ResNet* & 21.47(13.5) & 32.56(13.1) & 37.53(16.9) & 1157.17(740.4) & 33.24(15.6) & 5.34(3.2)\\
 DC$^{2}$-Net* & 19.06(13.0) & 30.64(17.1) & 33.06(15.2) & 1017.02(814.8) & 27.99(12.8) & 5.43(3.2)\\
 RecON* & 15.74(10.5) & 26.80(19.0) & 24.90(11.2) & 800.50(716.9) & 22.36(11.3) & 4.25(2.8)\\
 MoNet & 14.53(9.5) & 26.50(19.2) & 23.67(10.7) & 753.40(593.4) & 21.11(11.0) & 2.92(1.8)\\
 OSCNet & 14.17(9.5) & 25.42(19.0) & 23.25(10.5) & 714.22(526.5) & 20.62(10.5) & 2.69(1.7)\\
\cline{1-7}
 ReM$'$ & 13.13(10.3) & 24.07(16.0) & 20.79(10.7) & 599.77(526.4) & 18.21(10.1) & 4.60(2.1)\\
 ReM & \textcolor{black}{11.27(7.9)} & \textcolor{black}{19.94(10.9)} & \textcolor{black}{17.58(8.8)} & \textcolor{black}{497.85(389.0)} & \textcolor{black}{15.96(8.4)} & \textcolor{black}{4.38(2.0)}\\
 ReM+F$'$ & 10.13(6.6) & 18.18(9.1) & 16.41(8.4) & 459.95(366.2) & 14.55(7.6) & 3.60(1.8)\\
 ReM+F & \textcolor{black}{9.07(5.9)} & \textcolor{black}{17.27(8.7)} & \textcolor{black}{15.12(7.5)} & \textcolor{black}{421.35(316.3)} & \textcolor{black}{13.33(6.9)} & \textcolor{black}{3.25(1.6)}\\
 FiMA & \textcolor{blue}{8.61(5.9)} & \textcolor{blue}{16.16(8.1)} & \textcolor{blue}{13.78(6.4)} & \textcolor{blue}{391.77(298.4)} & \textcolor{blue}{12.39(6.2)} & \textcolor{blue}{2.09(1.2)}\\
\hline
\end{tabular}
\end{table}

The arm and carotid datasets were split into training/validation/test sets in a ratio of 375/104/104 and 276/78/78 scans, respectively, based on volunteer allocation. We performed random augmentations on each scan referring to ~\cite{luo2023multi}. We used the Adam optimizer to optimize the model. During the training phase, the epochs and batch size are set to 200 and 1, respectively. To avoid over-fitting, we set the initial learning rate to $2 \times 10^{-4}$ and used a learning rate decay strategy that halves the learning rate every 30 epochs. During the testing phase, the iteration epoch and learning rate are set to 60 and $2 \times 10^{-6}$, respectively.

\subsubsection{Quantitative and Qualitative Analysis.} 

\begin{figure}[t]
\centering
\includegraphics[width=\textwidth]{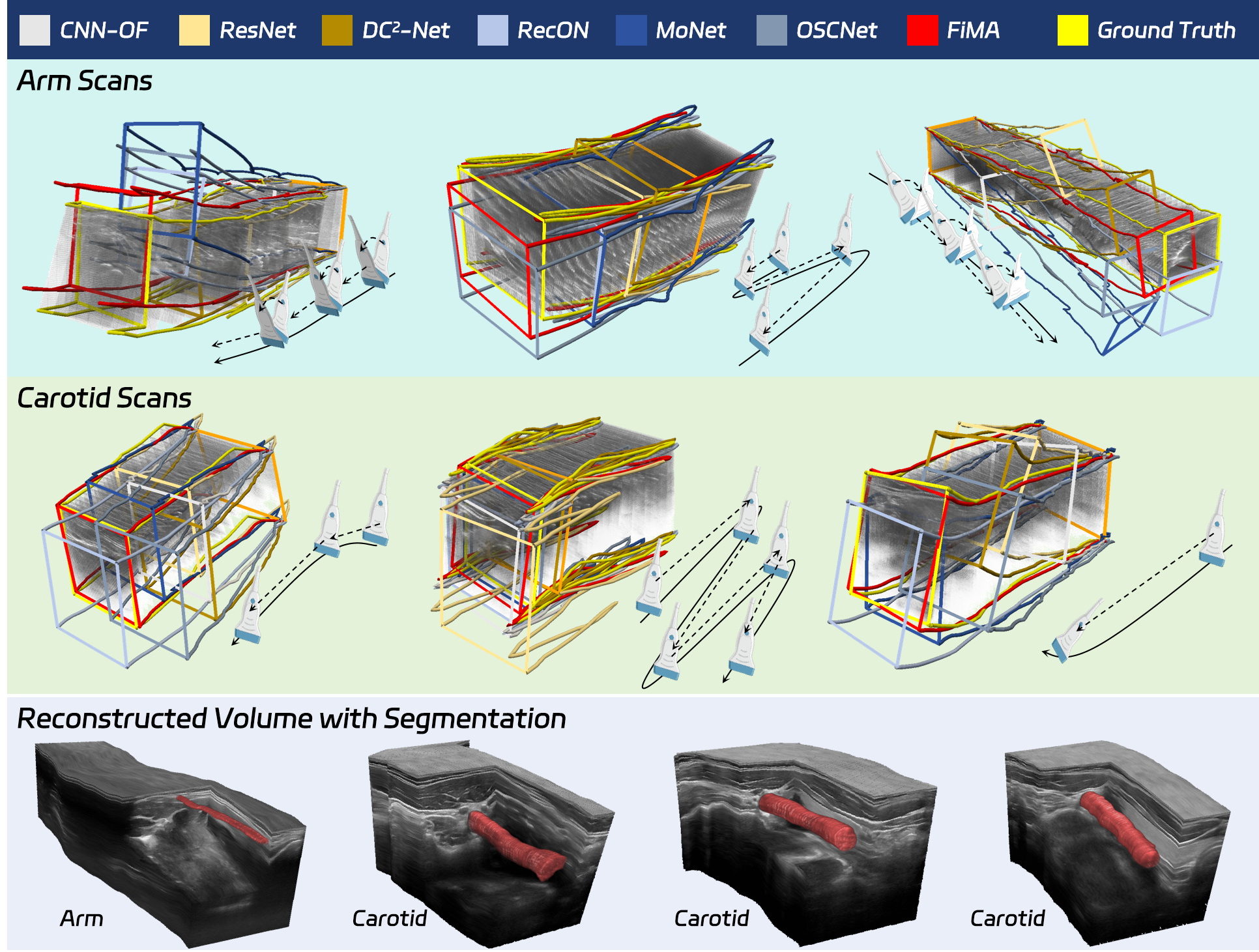}
\caption{Reconstruction examples produced by our proposed method. Red surface denotes the vessels reconstructed. Probe trajectory represents the scanning path.}
\label{fig:reconstruction-result}
\end{figure}

We use following metrics refering to~\cite{luo2023multi} to quantify the performance of FiMA: final drift rate (FDR), average drift rate (ADR), maximum drift (MD), sum of drift (SD), symmetric Hausdorff distance (HD), and mean error of angle (MEA). We compare FiMA with following methods: CNN-OF~\cite{prevost20183d}, ResNet~\cite{he2016deep}, DC$^{2}$-Net~\cite{guo2022ultrasound}, RecON~\cite{luo2023recon}, MoNet~\cite{luo2022deep}, OSCNet~\cite{luo2023multi}. All comparison methods were conducted following original experimental settings. The quantitative results are shown in Table~\ref{tab1}.

As seen in Table~\ref{tab1}, ReMamba outperforms single-IMU-based MoNet in several metrics, even in the absence of an IMU. Moreover, it demonstrates the effectiveness of our proposed multi-directional SSM, adaptive fusion strategy and online alignment strategy. The optimal result achieved further improvements over OSCNet, with 25.57\% /21.96\% and 39.24\%/36.43\% improvement in FDR/ADR on the arm and carotid datasets, respectively. FiMA achieves the state-of-the-art performance.


Figure~\ref{fig:reconstruction-result} displays representative reconstruction outcomes from all the methods of comparison. It is evident that our FiMA demonstrates superior performance and aligns with the ground truth more closely compared to other methods on both the arm and carotid datasets. Segmentation results on typical reconstructed volumes show that FiMA can reconstruct blood vessels well, which is expected to provide a reference for 3D analysis of anatomical structures.


\section{Conclusion}
In this study, we propose FiMA to exploits the efficient long-range dependency management capabilities of SSM. FiMA realises the capture of spatio-temporal information in multi-scale features, including these fine-grained features that are crucial in reconstruction. We innovate an multi-modal fusion strategy to adaptively extracted suitable information from multiple IMUs to guide reconstruction. We propose an online alignment strategy to ensure stable and accurate reconstruction performance of FiMA when inferring on unseen data. The experimental results on the arm and carotid datasets show that above methods have resulted in great performance gains and FiMA achieves state-of-the-art reconstruction performance.




%
%
%
\bibliographystyle{splncs04}
\bibliography{Paper-2611}
%
\end{document}